\documentclass[sigconf]{acmart}
\usepackage{algorithm}
\usepackage{algorithmic}

\usepackage{tikz}
\usepackage{graphicx}
\usetikzlibrary{positioning, fit, backgrounds, arrows.meta, calc, shapes.geometric}

\AtBeginDocument{%
  }

\copyrightyear{2026}
\acmYear{2026}
\setcopyright{none}
\acmDOI{}
\acmISBN{}
\acmConference[]{}{}{}
\acmBooktitle{}
\renewcommand\footnotetextcopyrightpermission[1]{}

\begin{document}
\setcounter{secnumdepth}{4}

\title{Connected Content Retriever: Dense Graph Edge Features for Pre-Ranking at LinkedIn }

  \settopmatter{authorsperrow=3}

  \author{Akhilesh Gupta}
  \authornote{These authors contributed equally to this research.}
  \affiliation{\institution{LinkedIn Corporation}\city{Mountain View}\state{CA}\country{USA}}
  \email{akgupta@linkedin.com}

  \author{Sudarshan Srinivasa Ramanujam}
  \authornotemark[1]
  \affiliation{\institution{LinkedIn Corporation}\city{Mountain View}\state{CA}\country{USA}}
  \email{sramanujam@linkedin.com}

  \author{Chirag Bhanuprasad Mehta}
  \authornotemark[1]
  \affiliation{\institution{LinkedIn Corporation}\city{Mountain View}\state{CA}\country{USA}}
  \email{chmehta@linkedin.com}

  \author{Reshma Asharaf Beena}
  \affiliation{\institution{LinkedIn Corporation}\city{Mountain View}\state{CA}\country{USA}}
  \email{rbeena@linkedin.com}

  \author{Dhritiman Das}
  \affiliation{\institution{LinkedIn Corporation}\city{Mountain View}\state{CA}\country{USA}}
  \email{dhdas@linkedin.com}

  \author{Birjodh Singh Tiwana}
  \affiliation{\institution{LinkedIn Corporation}\city{Mountain View}\state{CA}\country{USA}}
  \email{btiwana@linkedin.com}

  \author{Bhargavkumar Kanubhai Patel}
  \affiliation{\institution{LinkedIn Corporation}\city{Mountain View}\state{CA}\country{USA}}
  \email{bhapatel@linkedin.com}

  \author{Mack Lee}
  \affiliation{\institution{LinkedIn Corporation}\city{Mountain View}\state{CA}\country{USA}}
  \email{maclee@linkedin.com}
  
  \author{Renyi Tang}
  \affiliation{\institution{LinkedIn Corporation}\city{Mountain View}\state{CA}\country{USA}}
  \email{rtang@linkedin.com}

  \renewcommand{\shortauthors}{Akhilesh Gupta et al.}



  
  



\begin{abstract}
In large-scale recommendation systems like the LinkedIn Feed, content generated by a member's network (connections and follows) makes up over 70\% of impressions and engagement. It is therefore essential that the pre-ranking layer forwards the best possible few hundred candidates to the ranking layer. LinkedIn's professional knowledge graph carries engagement signals across both the first-degree network (connections and follows) and the second-degree network - posts that a 1st-degree connection reacted to, commented on, or reshared but did not author (a.k.a. stranger viral). Due to this fan-out, the resulting candidate index exceeds one billion; selection of activities from the viewer's network narrows it down to roughly tens of thousands of activities that must be scored within a $120$\,ms $p_{99}$ latency budget. We present Connected Content Retriever (CC Retriever), a pre-ranking system that scores these candidates with a full deep ranking model on GPUs at low latency. At its core is a sorted-search GPU primitive that joins dense graph-affinity features (viewer-to-author) with document-level features stored on the GPU at runtime in 5-10\,ms. The shift to GPU-served scoring enabled a 50$\times$ scale-up of the ranking model's parameters and delivered a +2.5\% lift in content time spent on the LinkedIn Feed in online experiments, significantly higher than the typical gains observed in LinkedIn Feed experiments. In this work, we describe the feature set we leverage from LinkedIn's economic graph and the model architecture used for scoring, with a particular emphasis on the online system that scales the stack. We present the join algorithm used to attach edge features (viewer-author features that capture pairwise affinity along the social graph, e.g., interaction recency and co-engagement) to documents, how request-level features are handled, and the GPU Retrieval-as-Ranking stack that now powers multiple use cases at LinkedIn. Ablations show that removing edge features causes a ~7.86\% drop in recall, underscoring the importance of graph-based connection signals that content embeddings alone were not able to capture.

\end{abstract}



\maketitle

\section{Introduction}
LinkedIn is the world's largest professional network, with hundreds of millions of monthly active members consuming posts from their professional graph: first-degree connections, follows, and second-degree posts propagated through engagement. Members produce on the order of hundreds of  million posts per day, but distribution is fundamentally graph-mediated rather than globally popularity-ranked: majority of Feed impressions are activities authored by, or engaged with by, an entity within the viewer's network neighborhood. Consequently, the relevance of a candidate post is jointly determined by (i) its semantic content and (ii) the viewer-to-content creator affinity along the social graph — a signal we term connection strength and treat as a first-class ranking feature, co-equal with topical interest. This contrasts with prior approaches that encode connection only as a graph-distance prior or apply it as a post-hoc reweighting of an otherwise content only ranker.

This work focuses on the pre-ranking (L1) stage of the Feed funnel, which is subject to a strict throughput-latency contract: reduce a candidate set of tens of thousand activities per request to a relevance ordered few hundreds subset, under a scoring latency budget of $120$\,ms $p_{99}$, while preserving sufficient fidelity in both the connection-strength and content-interest dimensions to avoid starving the downstream L2 ranker of discriminative signal. The central modeling challenge is therefore to jointly embed graph-affinity and content-semantic signals into a scoring function that is (a) cheap enough to evaluate over O($10^4$) candidates per request, and (b) calibrated enough that its top-K cutoff is approximately recall-preserving with respect to the L2 ranker's objective. We define an "\textbf{author}" as the LinkedIn member who originates a candidate item, whether by authoring an original post, resharing, reacting or commenting on another member's post.



In prior work \cite{ramanujam2026large} we showed that LLM based causal retrieval can be effectively scaled for the unconnected content surface, where item relevance is dominated by content semantics. However, the connected-content (CC) regime is qualitatively different: pairwise viewer–author edge features (tie strength, interaction recency, co-engagement, second-degree path statistics) carry a first-order signal that cannot be recovered from content embeddings alone. We empirically observed a statistically significant degradation in online engagement metrics in A/B tests when the CC pre-ranking layer was driven by LLM content embeddings without graph-edge conditioning; we report the corresponding offline ablation in the results section to quantify the contribution of edge features to top-K recall.

\begin{figure}
    \centering
    \includegraphics[width=0.9\linewidth]{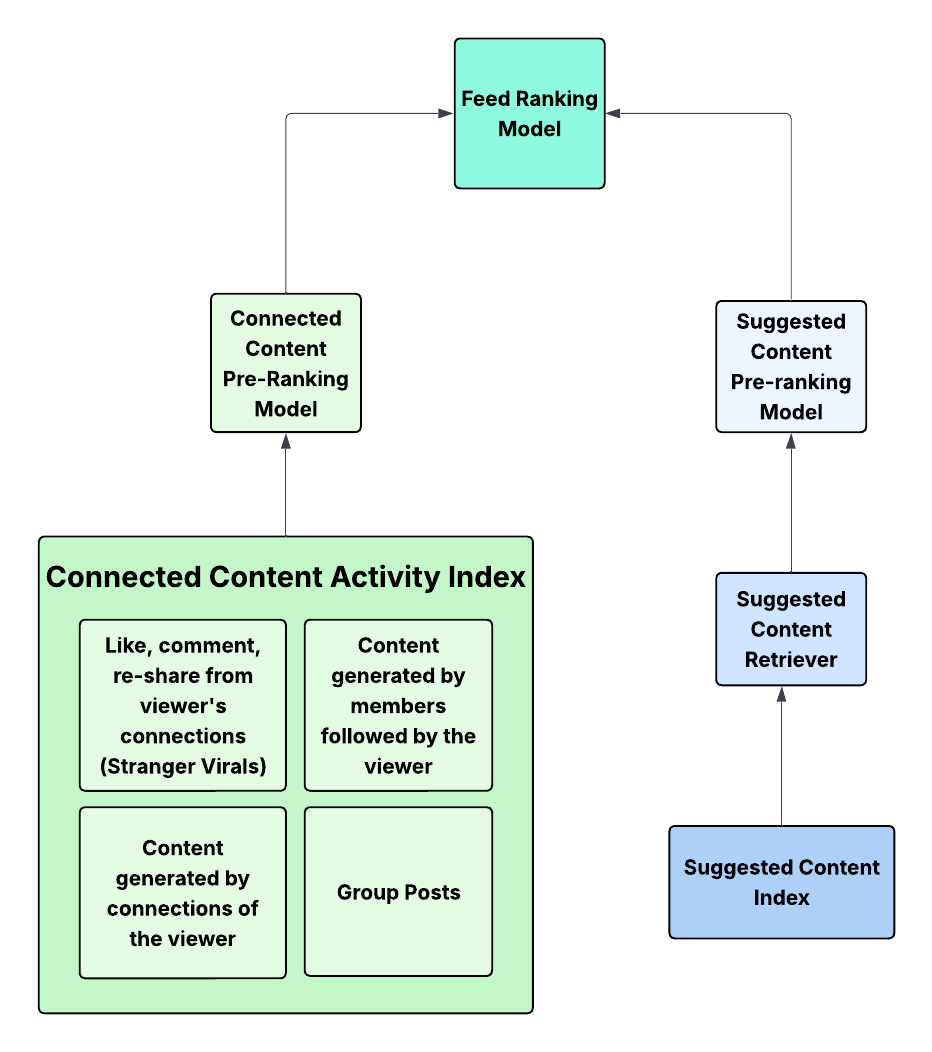}
    \caption{Feed Retrieval and Ranking System}
    \Description{Feed Retrieval and Ranking System}
    \label{fig:single_task_arch}
\end{figure}

The dominant constraint on this layer is its compute envelope: scoring on the order of tens of thousands of candidates per request under a P99 latency of 120 ms. Historically, this has forced the CC pre-ranker into a low-capacity regime — typically a shallow MLP or two-tower model with on the order of $O(10^5)$ trainable parameters — sized to amortize edge-feature lookup and per-candidate scoring within budget on CPU \cite{GhikeFollowfeed}. This capacity ceiling is the main bottleneck that prevents the layer from expressing higher-order interactions between the content and the graph signals.

The recent shift to GPU-resident scoring has enabled a class of retrieval-as-ranking (RAR) architectures in industrial recommender systems \cite{BorisyukCikm2024,zhai2023revisiting,Hong2025GESR}, in which a single model jointly performs candidate retrieval and fine-grained scoring on accelerator hardware. However, existing RAR formulations are predominantly two-tower or late-interaction designs optimized for content-only retrieval, and do not provide a tractable mechanism for fusing high-cardinality, pair-dependent edge features into the scoring function at the latency targets required for industrial pre-ranking — fundamentally because edge features are not factorizable into a static item-side representation. In this work we present an end-to-end redesign of the CC pre-ranking stack as a GPU-RAR system that (i) materializes viewer–candidate edge features inline on the GPU, (ii) fuses them with content embeddings through a non-factorizable interaction module, and (iii) sustains the tens of thousands candidates scoring at $120$\,ms $p_{99}$ contract while operating at a parameter scale several orders of magnitude beyond the prior CPU-bound baseline.



\section{Related Work}
LinkedIn's LiNR \cite{BorisyukCikm2024} pioneers live-updated, model-based retrieval with billion-scale indexes embedded in a differentiable GPU. This approach, named GPU Retrieval as Ranking (GPU-RAR), loads the entire candidate set into GPU memory and scores all candidates against the query in parallel via dense matrix operations, replacing approximate nearest-neighbor index lookups with exhaustive GPU search. SilverTorch \cite{xue2025silvertorch} extends this direction further. Instead of calling out to separate ANN and filtering services, it implements both as tensor operations: a fused int8 Inverted File Index kernel for nearest-neighbor search and a GPU bloom index that performs filtering with bitwise tensor operations. 
Both papers show that when retrieval runs on GPU, new opportunities emerge for richer scoring than two-tower decomposition allows.
 
Building on this model-based retrieval paradigm, Tencent \cite{lei2025efficient} deploys Wide \& Deep at retrieval scale by augmenting dual-tower embeddings with explicit sparse cross-feature interactions. These cross-features are binary indicators of whether a user has interacted with ads sharing a given attribute (categories, advertiser metadata, etc.), gated against real-valued aggregates such as click counts and CTRs over 1, 7, 15, and 30-day windows, and accelerated via a GPU-optimized compressed inverted list. While the system introduces explicit feature interaction into retrieval, the interaction module itself remains a sparse weighted aggregation operator rather than a general neural scorer operating over dynamically joined dense cross-features.
 
Beyond linear wide components, several systems pursue richer query-item interactions at retrieval scale through alternative architectural routes. SparCode \cite{su2023sparcode} enables all-to-all feature interactions between queries and items via an expressive parameterized scorer (e.g., MLPs, attention, or CrossNet), made tractable at retrieval time by quantizing queries into a small vocabulary of discrete codes so that the scorer can be evaluated offline over all (code, item) pairs, with the resulting scores cached in a sparse inverted index that is jointly trained end-to-end with the scorer and the codebook. HSNN \cite{rangadurai2024hsnn} introduces user-ad interaction features and an expressive non-linear scorer into retrieval while preserving sub-linear cost by jointly learning a hierarchical clustering of items together with the scorer, so that the costly interaction computation runs once per cluster rather than once per item. Both approaches escape the inner-product constraint of two-tower models by introducing an intermediate vocabulary, either codes over queries or clusters over items, that lets expensive scoring be amortized away from per-item online work, but they do not join explicit cross-features at serving time.
 
A related line of work focuses on richer query-item interactions at pre-ranking through architectural means. GESR \cite{Hong2025GESR} deploys a Mixture of Attention, combining hard-match attention, target-aware self-attention, and cross-attention, supported by custom kernels and serving caches. TARQ \cite{li2025tarq} brings target attention into pre-ranking at Taobao via residual quantization, with attention scores precomputed against a learned codebook. HIT \cite{yang2025hit} keeps two-tower decoupling intact and instead pregenerates holistic interaction vectors aligned at training time. Across these systems, the gains come from the modeling architectures that enable richer interactions. To fit within a millisecond-scale serving budget, techniques such as kernel-level optimization and residual quantization are employed.
 
Another line of work extends the cross-feature signal vocabulary available at serving time. InteractRank \cite{khandagale2025interactrank} augments a two-tower model at Pinterest with query-item statistics derived from two years of logs. AIF \cite{kou2025aif} performs user-side computation in parallel with retrieval and item-side computation in a nearline manner, with the residual user-item interaction approximated at serving time. In both cases, explicit cross-features enter pre-ranking through offline precomputation, with the resulting freshness characteristics depending on the update cadence.

\section{Features for Pre-Ranking Model}
\label{sec:features}

We use 3 different types of features for scoring our models. 

\begin{itemize}
    \item Document Features which is an attribute of candidate post (examples include item popularity, content embeddings of the post etc.)
    \item Request Features which are unique to a request (examples include viewer specific features like viewer embedding). 
    \item Edge features which need both viewer and author of each candidate post to derive the feature
\end{itemize}

Document features are indexed in GPU for scoring and request features are passed to the GPU at request time. Required attributes to construct the edge features are also passed to the GPU at request time and the edge feature is joined with the required documents. 

\subsection{Edge Feature Join At Scoring Time Online}

Edge features capture the \textit{relationship} between a viewer and content creator (author), independent of the specific post. 

\textbf{Viewer-Author Action Affinity} is an example of edge feature which encodes the viewer's historical engagement with each author across action types (like, comment, share, click, long-dwell, follow, etc.) and time buckets (last hour, last day, last week, etc.). This multi-dimensional vector captures both the intensity and the recency of engagement. Edge features like these are highly predictive since they provide strong information on the connection strength between the viewer and the author. However, joining them to the required documents at request time is computationally expensive given our latency cost. 


At serving time, we receive $N$ candidate documents, each with an author ID: $\mathbf{d} = [d_1, d_2, \ldots, d_N]$ and $K$ viewer-author affinity records $(\mathbf{a}, \mathbf{F})$ where $\mathbf{a} = [a_1, \ldots, a_K]$ are author IDs and $\mathbf{F} \in \mathbb{R}^{K \times D}$ are the corresponding feature vectors. We need to produce output features $\mathbf{O} \in \mathbb{R}^{N \times D}$ where:
\begin{equation}
    O_j = \begin{cases}
        F_i & \text{if } \exists i: a_i = d_j \\
        \mathbf{0} & \text{otherwise}
    \end{cases}
\end{equation}


 The naive approach would scan all $K$ affinity records for each candidate's author ID $d_j$ to find a match, yielding $O(N \cdot K)$ operations. With $N$ in the tens of thousands of candidate documents per request and $K$ on the order of a few thousand viewer-author affinity records , this product is in the order of $10^7$ which is prohibitive within our pre-ranking latency target.

  \begin{algorithm}[t]
  \caption{Edge Feature Join via Binary Search}
  \label{alg:join}
  \begin{algorithmic}[1]
  \REQUIRE Document author IDs $\mathbf{d} \in \mathbb{Z}^{B \times N}$
  \REQUIRE Affinity features $\mathbf{F} \in \mathbb{R}^{B \times K \times D}$
  \REQUIRE Affinity author IDs $\mathbf{a} \in \mathbb{Z}^{B \times K}$
  \ENSURE Output features $\mathbf{O} \in \mathbb{R}^{B \times N \times D}$

  \STATE \textbf{// Step 1: Sort affinity records by author ID}
  \STATE $\mathbf{a}_{sorted}, \mathbf{idx} \gets \text{sort}(\mathbf{a}, \text{dim}=1)$ \hfill $O(K \log K)$
  \STATE $\mathbf{F}_{sorted} \gets \text{gather}(\mathbf{F}, \mathbf{idx})$

  \STATE \textbf{// Step 2: Binary search for each document}
  \STATE $\mathbf{pos} \gets \text{searchsorted}(\mathbf{a}_{sorted}, \mathbf{d})$ \hfill $O(N \log K)$

  \STATE \textbf{// Step 3: Validate matches (clamp index to avoid OOB)}
  \STATE $\mathbf{pos}_{safe} \gets \min(\mathbf{pos},\; K{-}1)$
  \STATE $\mathbf{valid} \gets (\mathbf{pos} < K) \land (\mathbf{a}_{sorted}[\mathbf{pos}_{safe}] = \mathbf{d})$

  \STATE \textbf{// Step 4: Gather features for matches}
   \STATE $\mathbf{O} \gets \mathrm{Zeros}(B \times N \times D)$
  \STATE $\mathbf{O}[\mathbf{valid}] \gets \mathbf{F}_{sorted}[\mathbf{pos}_{safe}[\mathbf{valid}]]$

  \RETURN $\mathbf{O}$
  \end{algorithmic}
  \end{algorithm}


Algorithm \autoref{alg:join} leverages the fact that we can sort the viewer's affinity records once and then use binary search to find matches for all documents efficiently. 
  
  \textbf{Complexity analysis:}
  \begin{itemize}
      \item Sort: $O(K \log K)$
      \item Binary search: $O(N \log K)$
      \item \textbf{Total}: $O(K \log K + N \log K) = O\!\big((N+K)\log K\big)$, dominated by $N \log K$ when $N \gg K$ which is up to $200\times$ fewer operations than the $O(N \cdot K)$ naive scan
  \end{itemize}

\section{Modeling Architecture}
%
%

\label{sec:modeling}

This section describes the Connected Content (CC) pre-ranking model.
Given the three feature classes introduced in Section~3 - request
features $x_v$, document features $x_d$, and edge features $x_e$,
the model emits logits for three engagement objectives
(click, totalContributions (union of likes, comments and shares), long-dwell). Finally we use tunable weights (hyper-parameters) to compute the final score of the retriever model. 

\subsection{Architecture}
\label{sec:arch}

\begin{figure}[t]
  \centering
  \resizebox{\columnwidth}{!}{%
  \begin{tikzpicture}[
      font=\scriptsize,
      >={Latex[length=1.3mm]},
      line width=0.4pt,
      input/.style ={draw=blue!55!black,  rounded corners=1pt, fill=blue!7,
                     minimum width=1.45cm, minimum height=0.6cm, align=center, inner sep=1pt},
      derived/.style={draw=teal!70!black, rounded corners=1pt, fill=teal!10,
                     minimum width=1.5cm,  minimum height=0.6cm, align=center, inner sep=1pt},
      proc/.style  ={draw=gray!70!black,  rounded corners=1pt, fill=gray!15,
                     minimum width=1.6cm,  minimum height=0.6cm, align=center, inner sep=1pt},
      debias/.style={draw=orange!80!black,rounded corners=1pt, fill=orange!15,
                     minimum width=1.8cm,  minimum height=0.6cm, align=center, inner sep=1pt},
      head/.style  ={draw=green!55!black, rounded corners=1pt, fill=green!12,
                     minimum width=1.35cm, minimum height=0.5cm, align=center, inner sep=1pt},
      score/.style ={draw=green!60!black, rounded corners=1pt, fill=green!25,
                     minimum width=2.6cm,  minimum height=0.55cm, align=center, inner sep=1pt},
      a/.style     ={->, draw=gray!60!black, line width=0.4pt},
  ]
  \node[input] (mem)  at (0.725, 3.0) {Member\\emb.};
  \node[input] (item) at (2.325, 3.0) {Item\\emb.};
  \node[input] (cnt)  at (3.925, 3.0) {Item\\Features};
  \node[input] (sps)  at (5.525, 3.0) {Sparse\\features};
  \node[input] (edge) at (7.125, 3.0) {Edge features};

  \node[proc]    (tower) at (1.0, 1.9) {Emb.\ tower\\(MLP+BN)};
  \node[derived] (cos)   at (2.9, 1.9) {Chunked\\cos.\ sim.};

  \node[proc, minimum width=7.7cm] (cat) at (3.925,  0.85) {Concatenate};
  \node[proc, minimum width=7.7cm] (bb)  at (3.925,  0.05)
     {Backbone MLP (LayerNorm+ReLU+Dropout, multi-block)};

  \node[proc, minimum width=2.7cm] (logits) at (2.4, -0.95) {Task logits};
  \node[debias]                    (eb)     at (5.7, -0.95) {Position-debias};

  \node[head] (h1) at (0.95, -2.0) {$\sigma$ Click};
  \node[head] (h2) at (2.4,  -2.0) {$\sigma$ Contributions};
  \node[head] (h3) at (3.85, -2.0) {$\sigma$ Long-dwell};

  \node[score] (s) at (2.4, -2.95) {Multi-objective score $s(x)$};
  
  \draw[a] (mem.south)  -- (tower.north);
  \draw[a] (item.south) -- (tower.north);
  \draw[a] (mem.south)  to[bend right=8] (cos.north);
  \draw[a] (item.south) -- (cos.north);

  \draw[a] (tower.south) -- (tower.south |- cat.north);
  \draw[a] (cos.south)   -- (cos.south   |- cat.north);
  \draw[a] (cnt.south)   -- (cnt.south   |- cat.north);
  \draw[a] (sps.south)   -- (sps.south   |- cat.north);
  \draw[a] (edge.south)  -- (edge.south  |- cat.north);

  \draw[a] (cat) -- (bb);
  \draw[a] (bb)  -- (logits);
  \draw[a] (eb.west) -- (logits.east);

  \draw[a] (logits.south) -- (h1.north);
  \draw[a] (logits.south) -- (h2.north);
  \draw[a] (logits.south) -- (h3.north);
  \draw[a] (h1.south) -- (s.north);
  \draw[a] (h2.south) -- (s.north);
  \draw[a] (h3.south) -- (s.north);
  \end{tikzpicture}}%
  \caption{CC pre-ranking model architecture. Inputs are pretrained member and item embeddings, log-normalized count features, multi-hot sparse features, and edge features $x_e$ joined as dense tensors per Algorithm~\ref{alg:join}. Pretrained embeddings pass through a ReLU + batch-norm tower; a chunked cosine-similarity block is computed in the forward pass. All signals are concatenated and consumed by a backbone MLP whose logits are additively corrected by a Feed-position-debias embedding. Sigmoid task heads for click, total
  contributions and long-dwell are combined via tunable weights}
  \label{fig:cc_prerank}
  \end{figure}
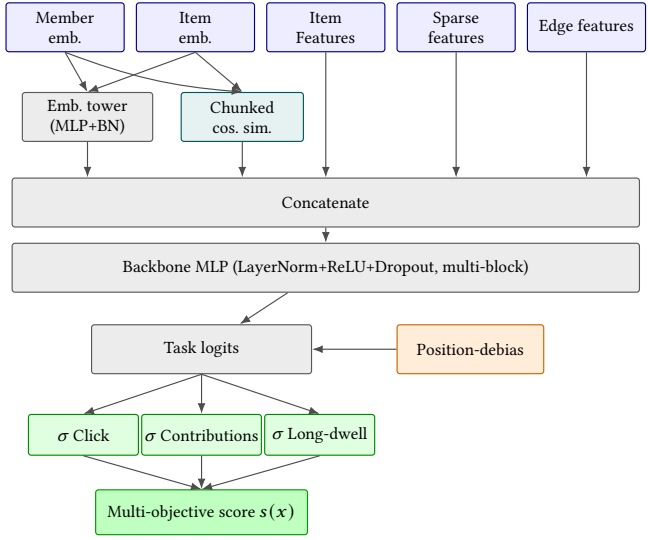

The CC pre-ranking model $f_\theta(x_v, x_d, x_e)$ is a multi-task
feed-forward network with $\sim O(10^{6})$ parameters, designed to score tens of thousands of candidates per request inside a 120\,ms p99 latency budget. The architecture is shown in \autoref{fig:cc_prerank}
the subsections below detail each block. Document features and edge features were already discussed in \autoref{sec:features} and we will not describe these again. 


\subsubsection{Embedding Tower}

The Pre-trained member (150-d from the top 3 medoids) and item (50-d) embeddings \cite{srinivasa2025linkedin} are
concatenated into a 200-d input and passed through a 2-layer ReLU + batch-norm sub-network. Batch-norm is used here so that running statistics provide a stable input scale to the downstream backbone.

\subsubsection{Derived Features}

 The member representation in \cite{srinivasa2025linkedin} is the concatenation of three 50-dimensional medoid embeddings; we treat each medoid as one 50-dim chunk aligned with the 50-d item embedding and compute a cosine similarity per chunk in float32, resulting in three scalars which are converted back to float16. 

Similarly we convert large integer count features to rate based features and apply log transformation bringing all of these to float16.


\subsubsection{Request Features}

Feature values for viewer specific features are passed along with the request. For the viewer side embeddings apart from the medoid based embeddings passed in the embedding tower, we also make use of the outputs of the final transformer layer of our sequence based ranking model which encodes information on all historical engagement of the member. \cite{hertel2026industrial}

 \subsubsection{Backbone MLP}

  The fused feature vector $x \in \mathbb{R}^{d_{\mathrm{in}}}$ flows
  through a stack of identical blocks of the form
  \begin{equation}
  h_{\ell+1} = \mathrm{Dropout}\!\big(\mathrm{LayerNorm}(\mathrm{ReLU}(W_\ell h_\ell + b_\ell))\big),
  \end{equation}
  followed by a final unactivated linear head producing the three task
  logits. Linear weights are Kaiming-normal initialized, biases are
  zero, and a small dropout rate is applied inside each hidden block.

  \subsubsection{Position-Debias Head}
  
  To prevent the backbone from learning ranker position bias, the Feed
  slot is treated as a \emph{debiasing covariate} rather than a feature:
  a learned embedding table $E_{\text{pos}}$ keyed by the (clipped) Feed
  slot is added to the backbone logits \cite{zhao2019recommending},
  \begin{equation}
  \hat{y} = \mathrm{Backbone}(x) + E_{\text{pos}}\!\big(\min(\text{slot},\,S_{\max})\big),
  \end{equation}
  where $S_{\max}$ is a cutoff above which slots are bucketed together.
  During training, a moderate dropout is applied to the position
  contribution; at inference and during validation the slot is clamped
  to a constant, zeroing out the bias term and yielding relevance-only
  logits.

\subsubsection{Multi-Objective Score}

At serving time the model emits logits for three engagement objectives:
  \begin{itemize} 
      \item \textbf{click}
      \item \textbf{totalContributions} --- union of likes, comments and shares
      \item \textbf{long-dwell}
  \end{itemize}
  The three predicted probabilities are combined via a linear combination with tunable per-objective weights (hyper-parameters) to produce the final retriever score.

\begin{equation}
s(x) = \sum_{k\in\{\text{click},\,\text{totalContributions},\,\text{long-dwell}\}} w_k\,\sigma(\hat{y}_k(x)).
\end{equation}

We also add an additional freshness boost to the final score based on age to encourage exploration in the system.

\section{Offline Evaluation}

\label{sec:eval}

We use recall@k for offline evaluation to check how well the pre-ranking model is aligned with the ranking model. The final ranked order of each session is taken as oracle. Candidates to the same session are scored with the retriever and we evaluate recall@10 to compare different variants of the offline retriever model. We chose 10 since it was most aligned with online metrics based on our empirical studies. 

 $$                                                                        \text{recall@}k = \frac{|R_K \cap M_k|}{K}. 
 $$                        
where $R_K$ is the ranker's top-$K$ activities (the relevant set) and $M_k$ is the retriever's top-$k$ activities.

\section{Online System}
\subsection{End-to-End System}
\label{sec:e2e-system}

This section describes the end-to-end online serving architecture at scale. The system scores tens of thousands of candidate documents per request using a learned model with dense embeddings, document features, and user-document edge features, all within the latency budget of a retrieval stage (${\sim}120$\,ms). We present a general architecture that decouples three concerns: (1)~candidate selection from a large corpus, (2)~feature storage and retrieval at corpus scale, and (3)~GPU-accelerated model scoring with batched inference. This separation enables independent scaling of each component and is applicable to any recommendation system where the retrieval stage must go beyond embedding
similarity to incorporate rich features.

\subsubsection{System Overview}
\label{sec:system-overview}

The online serving pipeline consists of three cooperating subsystems that
execute in sequence for each user request.

\begin{figure}[t]
  \centering
  \includegraphics[width=\columnwidth]{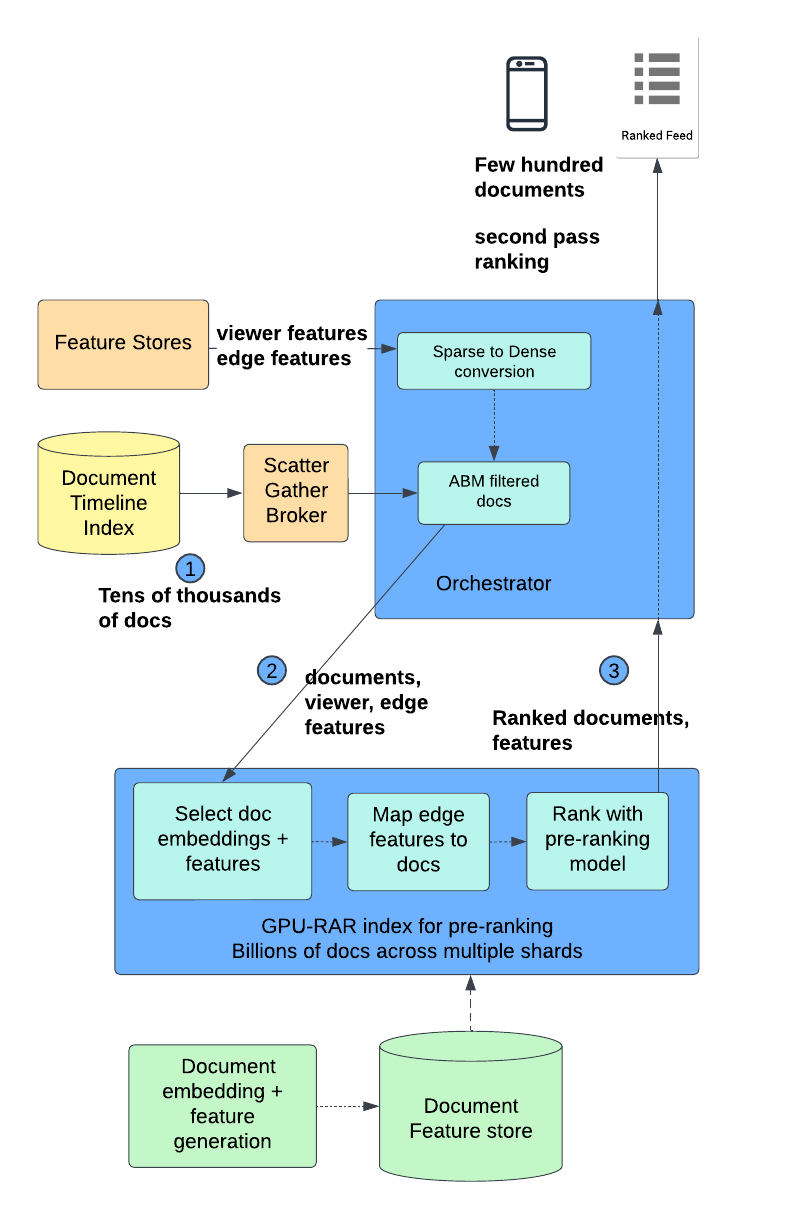}
  \caption{End-to-end architecture of the retrieval-as-ranking system.
  A user request flows through three stages: (1)~candidate selection from a
  graph-based index filters the corpus from billions to tens of thousand
  documents, (2)~the retrieval service generates request-level and edge
  features and assembles the scoring request, and (3)~the GPU inference
  server resolves candidate IDs to pre-indexed feature tensors, executes a
  batched model forward pass, and returns top-$K$ results to the ranking model.}
  \label{fig:e2e-architecture}
\end{figure}

\subsubsection{Stage 1: Candidate Selection via Graph-Based Filtering}
\label{sec:stage1-candidate-selection}

The first stage reduces the full document corpus ($N_{\text{corpus}}$) to a
tractable candidate set ($N_{\text{candidates}}$) for model scoring. In our
Feed recommendation scenario, the relevant document set for a user is defined
by a social graph: only documents authored or interacted with by entities
connected to the user are eligible. We use a \textbf{timeline index} that maintains the mapping from document authors to the timeline of their recent documents. This index supports
attribute-based matching (ABM): given a user, it returns all document IDs
whose author is within the user's network. This filtering operation reduces
the corpus by 4--5 orders of magnitude (from billions to tens of thousands).

\paragraph{Key design decisions.}
The timeline index uses a \textbf{scatter-gather} pattern across shards. The
index is partitioned (e.g., by document author ID), and an orchestration
broker fans out the user's network membership to the relevant shards, collects
matching document IDs, and deduplicates. This allows the index to scale
horizontally with corpus size.

\subsubsection{Stage 2: Retrieval Service (Orchestration \& Feature Assembly)}
\label{sec:stage2-retrieval-service}

The retrieval service is the orchestrator between candidate selection and GPU scoring. It receives the filtered candidate IDs from Stage~1, computes user-level and request-level features once per request, looks up the user-keyed edge affinity tensors (Section~\ref{sec:features}), and assembles a compact scoring payload comprising the candidate IDs, the user's query embedding, request-level features, and the edge affinity tensors keyed by entity ID. Crucially, \textbf{document features and document embeddings are not transmitted in the payload}; they are pre-indexed in the GPU inference server (Section~\ref{sec:stage3-gpu-server}) and resolved server-side by document ID, reducing per-request payload size by orders of magnitude.
\subsubsection{Stage 3: GPU Inference Server}
\label{sec:stage3-gpu-server}

The GPU inference server receives candidate IDs and user-side features from Stage~2, resolves them to pre-indexed feature tensors, and executes a batched model forward pass to produce the final ranking. Its architecture, including corpus-scale ingestion, ID resolution, batched scoring, and memory management is described in detail in Sections~\ref{sec:ingestion-pipeline} and~\ref{sec:gpu-rar}.

\subsection{Ingestion Pipeline: Keeping the Corpus Index Fresh}
\label{sec:ingestion-pipeline}

To keep the GPU index fresh against continuous document creation, update, and aging, we follow LiNR's offline-batch-plus-nearline-CDC ingestion pattern~\cite{BorisyukCikm2024}, consuming document updates as a stream and writing them to the GPU index in near-real-time. Our pipeline extends this with a two-class feature split: static features computed once at document creation (e.g., embeddings, content type), and dynamic features such as \texttt{activityPopularity} that evolve with member engagement. The two classes flow through separate streams so the index can refresh them independently rather than reprocessing the full feature set on every change. Static features arrive on an availability event triggered by document creation; any fields missing from that event are hydrated via gRPC calls to the corresponding feature services. Dynamic features are produced by a stateful streaming job that aggregates member interactions over fixed windows and emits periodic updates on its own stream. The ingestion job consumes both streams in parallel and writes updates independently to the index.

\paragraph{Storage efficiency at scale.}
At billions of documents, per-record layout choices have outsized impact on GPU memory. We extend LiNR's compact-integer attribute encoding~\cite{BorisyukCikm2024} to our categorical features using 32-bit integers, enabling vectorized GPU comparison during filtering and yielding an order-of-magnitude reduction over string representations. A daily background job evicts documents past the retention window, bounding the active corpus.

\paragraph{Scaling ingestion for a billion-document corpus.}
The ingestion subsystem maintains a GPU-resident index while processing up to tens of thousands of updates per second --- orders of magnitude above both the platform's post creation rate and LiNR's reported update rates~\cite{BorisyukCikm2024} --- with a maximum staleness target of a couple of minutes. Each post induces a long tail of downstream update events (engagement signals, attribute changes, reshares) that the index must continuously absorb to stay fresh. A CPU-resident parallel hash map $H: \text{DocID} \rightarrow \text{Position}$ (tens of GB for billions of entries) maps document IDs to their GPU tensor positions. Ingestion operates in two modes:

\paragraph{Bootstrap (server cold start).}
Bootstrap replays the CDC stream at cold start~\cite{BorisyukCikm2024}; we batch documents in groups of hundreds of thousands and apply a four-step per-batch protocol: (1)~parallel-identify new vs.\ existing document IDs via parallel iterator over $H$, (2)~sequential prefix-sum to assign storage positions for new documents, (3)~GPU scatter features to assigned positions, and (4)~update the hash map only after successful writes for transactional consistency. An earlier implementation used a tensor membership check (\texttt{isin}) for batch lookups; replacing it with the parallel hash-map iterator gave a $52\times$ speedup in average batch insert time.

\paragraph{Steady-state (serving + ingestion).}
Once serving begins, ingestion batch sizes are reduced by 3x to
avoid contending with model execution for GPU memory bandwidth. Parallel iterator performance keeps ingestion lag under $1$ minute.

\subsection{GPU Retrieval as Ranking Beyond Embedding Features}
\label{sec:gpu-rar}

  \begin{figure}[t]
  \centering
  \resizebox{\columnwidth}{!}{%
  \begin{tikzpicture}[
      font=\scriptsize,
      >=stealth,
      line width=0.3pt,
      box/.style={draw=blue!55!black, rounded corners=1pt, fill=blue!8,
                  minimum width=1.1cm, minimum height=0.8cm, align=center, inner sep=1pt},
      proc/.style={draw=gray!70!black, rounded corners=1pt, fill=gray!12,
                   minimum width=1.1cm, minimum height=0.8cm, align=center, inner sep=1pt},
      store/.style={draw=orange!75!black, rounded corners=1pt, fill=orange!12,
                    minimum width=1.6cm, minimum height=1.0cm, align=center, inner sep=1pt},
      output/.style={draw=green!55!black, rounded corners=1pt, fill=green!12,
                     minimum width=1.1cm, minimum height=0.8cm, align=center, inner sep=1pt},
      pathlbl/.style={font=\tiny\itshape, anchor=west},
      a/.style={->, draw=gray!60!black, line width=0.3pt},
  ]
  \node[store] (hix) at (1.5, 0)
      {Hash index\\\tiny DocID\,$\to$\,pos\\\tiny billions of entries};
  \node[store] (dfs) at (4.5, 0)
      {Doc feat.\ store\\\tiny billions of docs\\\tiny embs.\ + feats.};

  \node[box]  (evt)  at (0.0, 1.8) {Event\\stream\\\tiny 10s K/sec};
  \node[proc] (pbl)  at (1.5, 1.8) {Batch\\lookup};
  \node[proc] (psum) at (3.0, 1.8) {Prefix-\\sum};
  \node[proc] (scat) at (4.5, 1.8) {GPU\\scatter};

  \draw[a] (evt)        -- (pbl);
  \draw[a] (pbl)        -- (psum);
  \draw[a] (psum)       -- (scat);
  \draw[a] (pbl.south)  -- (hix.north);
  \draw[a] (scat.south) -- (dfs.north);

  \node[box]    (feed) at (0.0, -1.8) {Feed\\backend};
  \node[proc]   (cpu)  at (1.5, -1.8) {ID\\lookup};
  \node[proc]   (gth)  at (4.5, -1.8) {Feat.\\gather};
  \node[proc]   (mlp)  at (6.0, -1.8) {Scoring\\model};
  \node[output] (tk)   at (7.5, -1.8) {Top-$k$};

  \draw[a] (feed)      -- (cpu);
  \draw[a] (cpu)       -- (gth);
  \draw[a] (gth)       -- (mlp);
  \draw[a] (mlp)       -- (tk);
  \draw[a] (hix.south) -- (cpu.north);
  \draw[a] (dfs.south) -- (gth.north);

  \node[pathlbl] at (-0.55,  2.7) {(a) Write};
  \node[pathlbl] at (-0.55, -2.7) {(b) Read};
  \end{tikzpicture}}%
  \caption{Architecture of the GPU inference server. The write path (top) ingests events through a batch lookup, prefix-sum offsets, and a GPU scatter that updates the shared hash index and document feature
  store. The read path (bottom) resolves candidate document IDs against the hash index on the CPU, gathers features on the GPU, runs a batched scoring-model forward pass, and selects the top-$k$ documents.}
  \label{fig:gpu_server}
  \end{figure}
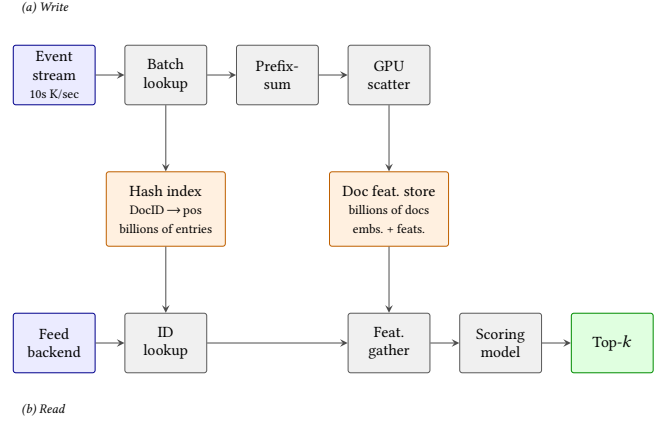

The GPU Retrieval-as-Ranking server is a Rust-based inference engine that maintains a real-time document index entirely in GPU memory. The Feed retrieval engine sends tens of thousands candidate ($p_{99}$) IDs per request; the inference server resolves them to feature tensors, assembles batched inputs, and executes a TorchScript model. The corpus of billions of documents are partitioned into multiple shards ($S$) (few hundred million docs in each). Each document stores a set of feature tensors $\{f_1, f_2, \ldots, f_k\}$---such as embedding vectors, activity signals, and content type encodings---with a combined per-document footprint of $D = \sum_{i=1}^{k} \text{sizeof}(f_i)$ bytes. Each shard must satisfy the memory constraint:
\begin{equation}
  \frac{N}{S} \times D \leq M_{\text{GPU}} - M_{\text{headroom}}
\end{equation}
where $M_{\text{GPU}}$ is the total device memory and $M_{\text{headroom}}$ reserves capacity for model weights, intermediate activations, and CUDA allocator overhead.

  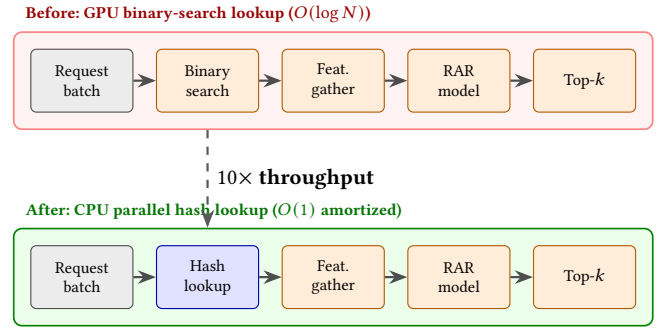
\begin{figure}[t]
  \centering
  \resizebox{\columnwidth}{!}{%
  \begin{tikzpicture}[
      >=Stealth,
      font=\scriptsize,
      line width=0.4pt,
      inputbox/.style={draw=gray!60!black, fill=gray!15, rounded corners=2pt,
                       minimum width=1.3cm, minimum height=0.8cm, align=center, inner sep=2pt},
      gpubox/.style={draw=orange!70!black, fill=orange!15, rounded corners=2pt,
                     minimum width=1.3cm, minimum height=0.8cm, align=center, inner sep=2pt},
      cpubox/.style={draw=blue!65!black, fill=blue!12, rounded corners=2pt,
                     minimum width=1.3cm, minimum height=0.8cm, align=center, inner sep=2pt},
      panel/.style={rounded corners=3pt, line width=0.6pt, inner sep=6pt},
      arr/.style={->, thick, draw=black!70},
      panellbl/.style={font=\scriptsize\bfseries, anchor=south west},
  ]

  \node[inputbox] (bin)  at (0.0,  1.5) {Request\\batch};
  \node[gpubox]   (bs)   at (1.6,  1.5) {Binary\\search};
  \node[gpubox]   (bfg)  at (3.2,  1.5) {Feat.\\gather};
  \node[gpubox]   (brar) at (4.8,  1.5) {RAR\\model};
  \node[gpubox]   (btk)  at (6.4,  1.5) {Top-$k$};

  \draw[arr] (bin)  -- (bs);
  \draw[arr] (bs)   -- (bfg);
  \draw[arr] (bfg)  -- (brar);
  \draw[arr] (brar) -- (btk);

  \begin{scope}[on background layer]
  \node[panel, draw=red!50, fill=red!5, fit=(bin)(btk)] (beforebox) {};
  \end{scope}
  \node[panellbl, text=red!60!black] at (beforebox.north west)
      {\ Before: GPU binary-search lookup ($O(\log N)$)};

  \node[inputbox] (ain)  at (0.0, -1.0) {Request\\batch};
  \node[cpubox]   (ac)   at (1.6, -1.0) {Hash\\lookup};
  \node[gpubox]   (afg)  at (3.2, -1.0) {Feat.\\gather};
  \node[gpubox]   (arar) at (4.8, -1.0) {RAR\\model};
  \node[gpubox]   (atk)  at (6.4, -1.0) {Top-$k$};

  \draw[arr] (ain)  -- (ac);
  \draw[arr] (ac)   -- (afg);
  \draw[arr] (afg)  -- (arar);
  \draw[arr] (arar) -- (atk);

  \begin{scope}[on background layer]
  \node[panel, draw=green!50!black, fill=green!8, fit=(ain)(atk)] (afterbox) {};
  \end{scope}
  \node[panellbl, text=green!45!black] at (afterbox.north west)
      {\ After: CPU parallel hash lookup ($O(1)$ amortized)};

  \draw[dashed, thick, ->, draw=black!65]
     (beforebox.south -| bs) -- node[right, font=\small\bfseries] {$10\times$ throughput}
     (afterbox.north -| ac);
  \end{tikzpicture}}%
  \caption{Effect of moving document ID-to-position resolution from GPU-side binary search to CPU-side parallel hash lookups. GPU binary search costs $O(\log N)$ per lookup and its latency scales with batch
  size; CPU parallel hash lookups are $O(1)$ amortized and stay roughly constant with batch size, yielding a $10\times$ throughput improvement on the inference path. Orange boxes are GPU operations, blue
  boxes CPU operations.}
  \label{fig:cpu-lookup-optimization}
  \end{figure}

\subsubsection{Scaling Inference: Reducing GPU Cost at Production Throughput}\leavevmode\\

Achieving few thousand QPS at $p_{99} \leq 120\text{ms}$ with a billions of documents GPU-resident index required minimizing GPU idle time and maximizing utilization per device. The fewer GPU cycles spent on non-inference work, the fewer shards and GPUs needed to meet throughput targets. We achieved this through three key optimizations: 
\begin{itemize}
    \item moving document ID resolution off the GPU
    \item batching variable-length requests into fixed-size tensors
    \item tuning the memory allocator to prevent fragmentation under sustained load
\end{itemize}

\paragraph{Document ID Resolution: GPU Kernel to CPU Parallel Lookup}\leavevmode\\

Each request carries tens of thousands of candidate IDs pre-filtered by upstream retrieval. The GPU inference server must resolve these to index positions using a hash map $H$ that maps each document ID to its storage location in the GPU index: for every candidate $q_i$, the server computes
\begin{equation}
  \text{pos}(q_i) = \begin{cases} H[q_i] & \text{if } q_i \in H \\ -1 & \text{otherwise} \end{cases}
\end{equation}
This resolution happens under a read lock on the index — a read-write lock at the index granularity. Each in-flight batch holds the read lock, while ingestion takes an exclusive write lock per upsert batch.

Resolving candidate IDs to index positions is structurally a hash-lookup problem, but a GPU-resident hash map is impractical for a streaming-update index: the ingestion rate (tens of thousands of upserts/sec, Section~\ref{sec:ingestion-pipeline}) would force frequent rebuilds, and the irregular, pointer-chasing access pattern is a poor match for GPU coalesced-memory execution. We therefore started with the standard workaround for keyed lookup on the GPU: store the ID$\rightarrow$position mapping as a sorted array in HBM and resolve each query with a custom binary-search kernel. For a single request, this completed in ${\sim}5\text{ms}$ over a few-hundred-million-entry index. However, once we added batching to improve GPU utilization, the kernel had to search $b\times$ tens of thousands of IDs against the same index, and its $O(\log N)$ per-lookup cost compounded with sequential HBM access. Profiling showed the kernel consumed ${>}50\%$ of model latency and scaled linearly with batch size (${\sim}5\text{ms}$ at batch 1 $\rightarrow$ ${\sim}47\text{ms}$ at batch 8), capping effective QPS.

Our solution was to move ID resolution off the GPU entirely, splitting work along the natural hardware boundary: irregular pointer-chasing on the CPU, dense tensor ops on the GPU. A CPU-resident parallel hash map resolves each ID in $O(1)$ amortized time and processes the full $b\times$ tens-of-thousands batch concurrently across CPU threads in ${\sim}1\text{ms}$, independent of batch size. This freed the GPU exclusively for feature gathering and model execution, yielding a $10\times$ throughput improvement. We also evaluated GPU-resident hash map implementations for $O(1)$ on-device lookups, but at our ingestion rate of tens of thousands of upserts/sec the hash map would require continuous rebuilds, negating its throughput advantage; for workloads with a near-static index this trade-off would flip.

\paragraph{Batching and Padding for Variable-Length Requests}\leavevmode\\

Requests carry variable candidate counts after position resolution. For a batch of $b$ requests with $n_i$ valid candidates each, we pad to $n_{\max} = \max(n_i)$ using a scatter-based approach:
\begin{enumerate}
  \item Compute scatter indices $(\text{batch\_idx}, \text{pos\_idx})$ from valid position masks---done once and reused for all features.
  \item Flatten valid positions across the batch and single-gather features from the GPU index.
  \item Scatter into the output tensor: $T[\text{batch\_idx}, :, \text{pos\_idx}] = \text{gathered\_features}$.
\end{enumerate}

\noindent\textit{Example with $b = 2$:} $R_0$ has valid positions $[10, 42, 87]$, $R_1$ has $[5, 63]$.

\begin{enumerate}
  \item Scatter indices: $R_0 \rightarrow (0,0), (0,1), (0,2)$;\; $R_1 \rightarrow (1,0), (1,1)$
  \item Flatten positions $\rightarrow [10, 42, 87, 5, 63]$, single gather from GPU index
  \item Scatter into $T[b{=}2,\; n_{\max}{=}3,\; \text{features}]$:
\end{enumerate}
\begin{align*}
  T[0,:] &= [\text{features}[10],\; \text{features}[42],\; \text{features}[87]] \\
  T[1,:] &= [\text{features}[5],\; \text{features}[63],\; \texttt{pad}]
\end{align*}

Padded positions can produce non-trivial MLP scores, so we mask them out with  $\text{scores}[i,j] = -\infty$.

\paragraph{GPU Memory Tuning Under Constrained Headroom}\leavevmode\\

With few hundred million document tensors consuming the majority of GPU memory per shard, the remaining headroom for model activations is tightly constrained:
\begin{equation}
  M_{\text{total}} = M_{\text{index}} + M_{\text{model}} + M_{\text{activation}} + M_{\text{allocator}}
\end{equation}
During serving, each batch requires the model to score $b \times n_{\max}$ candidates, where $b$ is the batch size and $n_{\max}$ is the count of candidates padded per request. For a model with few million parameters, each forward pass allocates intermediate activation tensors proportional to $b \times n_{\max} \times d_{\text{hidden}}$ across both layers, plus temporary tensors for feature gathering, embedding dot products, and affinity lookups. Batching and high volume of documents to score creates significant transient memory pressure on top of the document index already occupying the majority of GPU memory.

Because the document index occupies the majority of HBM per shard, only a few hundred MB of contiguous headroom remain for the forward pass. Under variable-batch load, this headroom fragmented and produced two distinct failures: (i) classic fragmentation OOM, and (ii) \texttt{CUBLAS\_STATUS\_ALLOC\_FAILED}, where cuBLAS could not claim a workspace from the fragmented headroom mid-forward. We address them independently. Setting the PyTorch garbage-collection threshold to $0.6$ reclaims unused cached blocks once $60\%$ of reserved memory is occupied, bounding fragmentation across batch sizes. We then pre-allocate a fixed cuBLAS workspace pool of $10\times 4096\,\text{KB}$ ($\sim$40 MB): $4096\,\text{KB}$ is the smallest per-workspace size at which cuBLAS still selects the optimal Hopper matmul kernel, and depth $10$ matches in-flight CUDA stream concurrency at $p_{99}$, eliminating per-stream workspace contention.

\section{Results}

\subsection{GPU-RAR Benchmarking Results}

\autoref{tab:benchmark} shows the effect of each optimization on the QPS handled. Moving the lookups to the CPU had the biggest effect. 
\begin{table}[h!]
  \centering
  \caption{Throughput progression of the GPU inference server under a 120ms $p_{99}$ SLA. Each row adds one optimization to the previous configuration. Batch size denotes concurrent requests scored in a single GPU forward pass.}
  \label{tab:benchmark}
  \footnotesize
  \resizebox{\columnwidth}{!}{%
  \begin{tabular}{clcc}
    \toprule
    \textbf{Stage} & \textbf{Configuration} & \textbf{Batch Size} & \textbf{QPS} \\
    \midrule
    0 & Baseline: single-request MLP forward & b & OOM \\
    1 & \quad + memory tuning + pre-materialized tensors & b & x \\
    2 & \quad + multi-request batching & $\leq 16b$ & 1.75x \\
    3 & \quad + CPU parallel hash map lookups & 32b & \textbf{10x} \\
    \bottomrule
  \end{tabular}
  }
\end{table}

\subsection{Modeling Results}

The following section illustrates the AB test results when a trained model with all the features and infra optimizations discussed were ramped online against the older model and system.  

\begin{itemize}
    \item Content Time Spent in LinkedIn up by \textbf{2.5\%} 
    \item Increase in volume of sessions by \textbf{0.1\%}
    \item Reduction in skipped updates by \textbf{0.8\%}
    \item Increase in professional interaction in the platform (active and passive engagement combined) by \textbf{2.68\%}
\end{itemize}

All results cited above have p-values < 0.0001 and are statistically significant. 

\subsubsection{Ablation Studies on Features}
  \begin{table}[H]
  \centering
  \small
  \setlength{\tabcolsep}{6pt}
  \caption{Feature ablation on the CC pre-ranker. Each row disables one input stream while holding architecture and training identical to the full model. Recall@$10$ is reported as a percent change relative
  to the full model on a fixed offline evaluation dataset.}
  \label{tab:feature_ablation}
  \begin{tabular}{lc}
  \toprule
  \textbf{Variant} & \textbf{Recall@10 $\Delta$} \\
  \midrule
  Full model (this work)               & ---            \\
  \quad w/o edge features              & $-7.86\%$      \\
  \quad w/o content features           & $-3.05\%$       \\
  \quad w/o item features              & $-2.5\%$       \\
  \bottomrule
  \end{tabular}
  \end{table}

\autoref{tab:feature_ablation} shows the impact that various features have on the recall metric. We can clearly see that the edge features have the heaviest impact when removed in the model on top of LLM based content features and other item features. 

  \subsubsection{Parameter-Scaling Ablation}
  \label{sec:parameter-scale}

  To verify that the deployed few-million parameter backbone is the right operating point, we ran a controlled scaling sweep that holds the feature set, training recipe, and embedding tower fixed and co-scales backbone width and depth across two orders of magnitude in parameter count. Offline retrieval quality saturates quickly past the anchor: scaling nearly two orders of magnitude beyond the deployed size yields only
  $+0.20\%$ Recall@$10$, with marginal returns falling below $0.01\%$ per doubling at the upper end (Table~\ref{tab:param_scale_joint}). The shape matches diminishing-returns scaling laws reported for tabular
   and CTR models.

  \begin{table}[t]
  \centering
  \small
  \setlength{\tabcolsep}{6pt}
  \caption{Joint parameter-scaling tradeoff. Backbones are reported as parameter-count multipliers of the deployed anchor ($1\times$). $\Delta$Recall@$10$ is relative to the anchor; relative QPS is
  per-replica throughput at a $120$\,ms $p_{99}$ SLA, normalized to the smallest measured variant ($1.00\times$). Offline quality saturates quickly while throughput degrades nearly linearly with parameter
  count.}
  \label{tab:param_scale_joint}
  \begin{tabular}{lrr}
  \toprule
  \textbf{Backbone (vs.\ deployed)} & \textbf{$\Delta$Recall@$10$} & \textbf{Rel.\ QPS} \\
  \midrule
  Deployed ($1\times$)   & ---        & deployed     \\
  $\sim$$4\times$        & $+0.11\%$  & $0.93\times$ \\
  $\sim$$8\times$        & $+0.15\%$  & $0.89\times$ \\
  $\sim$$15\times$       & $+0.16\%$  & $0.78\times$ \\
  $\sim$$40\times$       & $+0.20\%$  & $0.67\times$ \\
  $\sim$$75\times$       & $+0.20\%$  & $0.56\times$ \\
  $\sim$$400\times$      & N/A        & OOM \\
  \bottomrule
  \end{tabular}
  \end{table}

  Online throughput pays a steep and roughly linear cost for those negligible offline gains. Under a $120$\,ms $p_{99}$ SLA, each doubling of the backbone costs roughly $10$--$15\%$ of sustainable
  per-replica QPS through the mid-range; near two orders of magnitude beyond the deployed size, sustainable QPS halves; beyond that, throughput collapses further, and the largest variant we attempted fails
  to load on a single GPU due to memory pressure. The deployed operating point therefore was kept at $\sim O(10^{6})$ for this tradeoff.

\section{Conclusion}
In this work we showed how we completely overhauled the connected content retrieval and pre-ranking stack in LinkedIn. We were able to move to GPU based pre-ranking where we could score larger models with complex edge based features for a large document set under $120$\,ms $p_{99}$ latency. We shared the end to end infrastructure and all the optimizations done to ship this system online which is now powering LinkedIn Feed. We have only just begun in this domain. Following this ground up rebuild, we intend to try more complex modeling architectures at the pre-ranking layer and are looking into sequence modeling. In parallel we are also working on leveraging more powerful language models to encode topical relevance along with the strong graph based features from the LinkedIn economic graph to power the Feed.

\section*{Acknowledgements}

This work represents the joint efforts across multiple teams in LinkedIn without whom this would not have been possible. We would like to thank (in alphabetical order) Adam Tang, Aman Mehrotra, Amithesh Ramesh, Amy Yang, Andrei Akterskii, Angela Shao, Anita Savadi, Antonio Alonso, Aslan (Mu) Bai, Bef Ayenew, Bo Wang, Christine Choi, Deepak Manoharan, Dru Pollini, Faizaan Charania, Fenil Fadadu, Hemeng Tao, Hristo Danchev, Jeffrey Zhao, Manika Agarwal, Mehiar Dabbagh, Mina Doroud, Mohit Kothari, Nick Ghotbi, Parminder Gill, Ping Jin, Pratik Dixit, Rahul Raja, Rajdeep Das, Rajeev Kumar, Rakesh Chalasani, Ronak Kaoshik, Russell Chan, Samaneh Moghaddam, Satyajeet Kumar, Saurabh Kataria, Shirley Jiang, Siddharth Dangi, Steve Jan, Swapnil Rajaram Patil, Tim Chao, Tim Jurka, Tugrul Bingol, Vibhuti Sengar, Vishal Shah, Wei Zhou, Yi Shen, Ying Xuan, Youngchae Kim, Yubo Ouyang, and Yuyang Tian for contributing to and supporting this work.

In preparing this paper, we used Claude Opus 4.6 (Anthropic) to identify and correct typos, grammatical errors, and collaboratively generate flow diagrams presented in the paper. Algorithm \autoref{alg:join} was also generated by Claude to convert production code to pseudocode. We acknowledge the contributions of Claude in enhancing the writing process while maintaining full academic and professional integrity.

 No AI tools were used for experimental design, model training, hyperparameter selection, statistical analysis, A/B test execution, or the formulation of scientific conclusions. All offline metrics, online A/B test results, and throughput benchmarks reported in this paper were produced by the authors using LinkedIn's production infrastructure. The authors reviewed, validated, and approved all AI-generated content and retain full responsibility for the scientific accuracy and integrity of this work.


\bibliographystyle{ACM-Reference-Format}
\balance
\bibliography{bibliography}


\begin{thebibliography}{16}


\ifx \showCODEN    \undefined \def \showCODEN     #1{\unskip}     \fi
\ifx \showISBNx    \undefined \def \showISBNx     #1{\unskip}     \fi
\ifx \showISBNxiii \undefined \def \showISBNxiii  #1{\unskip}     \fi
\ifx \showISSN     \undefined \def \showISSN      #1{\unskip}     \fi
\ifx \showLCCN     \undefined \def \showLCCN      #1{\unskip}     \fi
\ifx \shownote     \undefined \def \shownote      #1{#1}          \fi
\ifx \showarticletitle \undefined \def \showarticletitle #1{#1}   \fi
\ifx \showURL      \undefined \def \showURL       {\relax}        \fi
\providecommand\bibfield[2]{#2}
\providecommand\bibinfo[2]{#2}
\providecommand\natexlab[1]{#1}
\providecommand\showeprint[2][]{arXiv:#2}

\bibitem[Borisyuk et~al\mbox{.}(2024)]%
        {BorisyukCikm2024}
\bibfield{author}{\bibinfo{person}{Fedor Borisyuk}, \bibinfo{person}{Qingquan
  Song}, \bibinfo{person}{Mingzhou Zhou}, \bibinfo{person}{Ganesh
  Parameswaran}, \bibinfo{person}{Madhu Arun}, \bibinfo{person}{Siva Popuri},
  \bibinfo{person}{Tugrul Bingol}, \bibinfo{person}{Zhuotao Pei},
  \bibinfo{person}{Kuang-Hsuan Lee}, \bibinfo{person}{Lu Zheng},
  \bibinfo{person}{Qizhan Shao}, \bibinfo{person}{Ali Naqvi},
  \bibinfo{person}{Sen Zhou}, {and} \bibinfo{person}{Aman Gupta}.}
  \bibinfo{year}{2024}\natexlab{}.
\newblock \showarticletitle{{LiNR}: Model Based Neural Retrieval on {GPUs} at
  {LinkedIn}}. In \bibinfo{booktitle}{\emph{Proceedings of the 33rd ACM
  International Conference on Information and Knowledge Management (CIKM
  '24)}}. \bibinfo{publisher}{ACM}, \bibinfo{address}{New York, NY, USA},
  \bibinfo{pages}{1--8}.
\newblock
\href{https://doi.org/10.1145/3627673.3680091}{doi:\nolinkurl{10.1145/3627673.3680091}}


\bibitem[Ghike and Gupta(2016)]%
        {GhikeFollowfeed}
\bibfield{author}{\bibinfo{person}{Swapnil Ghike} {and}
  \bibinfo{person}{Shubham Gupta}.} \bibinfo{year}{2016}\natexlab{}.
\newblock \bibinfo{title}{{FollowFeed}: {LinkedIn}'s Feed Made Faster and
  Smarter}.
\newblock \bibinfo{howpublished}{LinkedIn Engineering Blog}.
\newblock
\urldef\tempurl%
\url{https://www.linkedin.com/blog/engineering/feed/followfeed-linkedin-s-feed-made-faster-and-smarter}
\showURL{%
\tempurl}
\newblock
\shownote{Accessed: 2026-04-30}.


\bibitem[Hertel et~al\mbox{.}(2026)]%
        {hertel2026industrial}
\bibfield{author}{\bibinfo{person}{Lars Hertel}, \bibinfo{person}{Gaurav
  Srivastava}, \bibinfo{person}{Syed~Ali Naqvi}, \bibinfo{person}{Satyam
  Kumar}, \bibinfo{person}{Yue Zhang}, \bibinfo{person}{Borja Ocejo},
  \bibinfo{person}{Benjamin Zelditch}, \bibinfo{person}{Adrian Englhardt},
  \bibinfo{person}{Hailing Cheng}, \bibinfo{person}{Andy Hu}, {et~al\mbox{.}}}
  \bibinfo{year}{2026}\natexlab{}.
\newblock \showarticletitle{An Industrial-Scale Sequential Recommender for
  LinkedIn Feed Ranking}.
\newblock \bibinfo{journal}{\emph{arXiv preprint arXiv:2602.12354}}
  (\bibinfo{year}{2026}).
\newblock


\bibitem[Hong et~al\mbox{.}(2025)]%
        {Hong2025GESR}
\bibfield{author}{\bibinfo{person}{Juhee Hong}, \bibinfo{person}{Meng Liu},
  \bibinfo{person}{Shengzhi Wang}, \bibinfo{person}{Xiaoheng Mao},
  \bibinfo{person}{Huihui Cheng}, \bibinfo{person}{Leon Gao},
  \bibinfo{person}{Christopher Leung}, \bibinfo{person}{Jin Zhou},
  \bibinfo{person}{Chandra~Mouli Sekar}, \bibinfo{person}{Zhao Zhu},
  \bibinfo{person}{Ruochen Liu}, \bibinfo{person}{Tuan Trieu},
  \bibinfo{person}{Dawei Sun}, \bibinfo{person}{Jeet Kanjani},
  \bibinfo{person}{Rui Li}, \bibinfo{person}{Jing Qian}, \bibinfo{person}{Xuan
  Cao}, \bibinfo{person}{Minjie Fan}, {and} \bibinfo{person}{Mingze Gao}.}
  \bibinfo{year}{2025}\natexlab{}.
\newblock \bibinfo{title}{Generative Early Stage Ranking}.
\newblock
\showeprint[arxiv]{2511.21095}


\bibitem[Khandagale et~al\mbox{.}(2025)]%
        {khandagale2025interactrank}
\bibfield{author}{\bibinfo{person}{Sujay Khandagale}, \bibinfo{person}{Bhawna
  Juneja}, \bibinfo{person}{Prabhat Agarwal}, \bibinfo{person}{Aditya
  Subramanian}, \bibinfo{person}{Jaewon Yang}, {and} \bibinfo{person}{Yuting
  Wang}.} \bibinfo{year}{2025}\natexlab{}.
\newblock \showarticletitle{InteractRank: Personalized Web-Scale Search
  Pre-Ranking with Cross Interaction Features}. In
  \bibinfo{booktitle}{\emph{Companion Proceedings of the ACM Web Conference
  2025 (WWW Companion '25)}}. \bibinfo{publisher}{ACM}, \bibinfo{address}{New
  York, NY, USA}, \bibinfo{pages}{1--9}.
\newblock
\href{https://doi.org/10.1145/3701716.3715239}{doi:\nolinkurl{10.1145/3701716.3715239}}


\bibitem[Kou et~al\mbox{.}(2025)]%
        {kou2025aif}
\bibfield{author}{\bibinfo{person}{Zhi Kou}, \bibinfo{person}{Xiang-Rong
  Sheng}, \bibinfo{person}{Shuguang Han}, \bibinfo{person}{Zhishan Zhao},
  \bibinfo{person}{Yueyao Cheng}, \bibinfo{person}{Han Zhu},
  \bibinfo{person}{Jian Xu}, {and} \bibinfo{person}{Bo Zheng}.}
  \bibinfo{year}{2025}\natexlab{}.
\newblock \bibinfo{title}{AIF: Asynchronous Inference Framework for
  Cost-Effective Pre-Ranking}.
\newblock
\showeprint[arxiv]{2511.12934}


\bibitem[Lei et~al\mbox{.}(2025)]%
        {lei2025efficient}
\bibfield{author}{\bibinfo{person}{Yifan Lei}, \bibinfo{person}{Jiahua Luo},
  \bibinfo{person}{Tingyu Jiang}, \bibinfo{person}{Bo Zhang},
  \bibinfo{person}{Lifeng Wang}, \bibinfo{person}{Dapeng Liu},
  \bibinfo{person}{Zhaoren Wu}, \bibinfo{person}{Haijie Gu},
  \bibinfo{person}{Huan Yu}, {and} \bibinfo{person}{Jie Jiang}.}
  \bibinfo{year}{2025}\natexlab{}.
\newblock \bibinfo{title}{An Efficient Embedding Based Ad Retrieval with
  {GPU}-Powered Feature Interaction}.
\newblock
\showeprint[arxiv]{2511.22460}


\bibitem[Li et~al\mbox{.}(2025)]%
        {li2025tarq}
\bibfield{author}{\bibinfo{person}{Yutong Li}, \bibinfo{person}{Yu Zhu},
  \bibinfo{person}{Yichen Qiao}, \bibinfo{person}{Ziyu Guan},
  \bibinfo{person}{Lv Shao}, \bibinfo{person}{Tong Liu}, {and}
  \bibinfo{person}{Bo Zheng}.} \bibinfo{year}{2025}\natexlab{}.
\newblock \bibinfo{title}{Equip Pre-ranking with Target Attention by Residual
  Quantization}.
\newblock
\showeprint[arxiv]{2509.16931}


\bibitem[Ramanujam et~al\mbox{.}(2026)]%
        {ramanujam2026large}
\bibfield{author}{\bibinfo{person}{Sudarshan~Srinivasa Ramanujam},
  \bibinfo{person}{Antonio Alonso}, \bibinfo{person}{Saurabh Kataria},
  \bibinfo{person}{Siddharth Dangi}, \bibinfo{person}{Akhilesh Gupta},
  \bibinfo{person}{Birjodh~Singh Tiwana}, \bibinfo{person}{Manas~Haribhai
  Somaiya}, \bibinfo{person}{Luke Simon}, \bibinfo{person}{David Byrne},
  \bibinfo{person}{Sojeong Ha}, \bibinfo{person}{Sen Zhou},
  \bibinfo{person}{Andrei Akterskii}, \bibinfo{person}{Zhanglong Liu},
  \bibinfo{person}{Samira Sriram}, \bibinfo{person}{Zihan Xiong},
  \bibinfo{person}{Zhoutao Pei}, \bibinfo{person}{Angela Shao},
  \bibinfo{person}{Alex Li}, \bibinfo{person}{Annie Xiao},
  \bibinfo{person}{Caitlin Kolb}, \bibinfo{person}{Thomas Kistler},
  \bibinfo{person}{Zach Moore}, {and} \bibinfo{person}{Hamed Firooz}.}
  \bibinfo{year}{2026}\natexlab{}.
\newblock \showarticletitle{Large Scale Retrieval for the {LinkedIn} Feed Using
  Causal Language Models}. In \bibinfo{booktitle}{\emph{Proceedings of the AAAI
  Conference on Artificial Intelligence}}, Vol.~\bibinfo{volume}{40}.
  \bibinfo{publisher}{AAAI Press}, \bibinfo{address}{Washington, DC, USA},
  \bibinfo{pages}{40101--40109}.
\newblock


\bibitem[Rangadurai et~al\mbox{.}(2024)]%
        {rangadurai2024hsnn}
\bibfield{author}{\bibinfo{person}{Kaushik Rangadurai}, \bibinfo{person}{Siyang
  Yuan}, \bibinfo{person}{Minhui Huang}, \bibinfo{person}{Yiqun Liu},
  \bibinfo{person}{Golnaz~Ghasemiesfeh Ahsan}, \bibinfo{person}{Yunzhong Tian},
  \bibinfo{person}{Siyu Lu}, \bibinfo{person}{Xilun Zheng},
  \bibinfo{person}{Ming Hu}, \bibinfo{person}{Haixin Liang},
  \bibinfo{person}{Qingwen Li}, \bibinfo{person}{Fedor Cao},
  \bibinfo{person}{Xiaoqiang Liu}, {and} \bibinfo{person}{Rong-En Sun}.}
  \bibinfo{year}{2024}\natexlab{}.
\newblock \bibinfo{title}{Hierarchical Structured Neural Network for
  Retrieval}.
\newblock
\showeprint[arxiv]{2408.06653}


\bibitem[Srinivasa~Ramanujam et~al\mbox{.}(2025)]%
        {srinivasa2025linkedin}
\bibfield{author}{\bibinfo{person}{Sudarshan Srinivasa~Ramanujam},
  \bibinfo{person}{Akanksha Bindal}, \bibinfo{person}{Yu Jiang},
  \bibinfo{person}{Timothy~J Hazen}, \bibinfo{person}{David Golland},
  \bibinfo{person}{Fengyu Zhang}, \bibinfo{person}{Daqi Sun},
  \bibinfo{person}{Wanning Li}, \bibinfo{person}{Birjodh~Singh Tiwana},
  \bibinfo{person}{Siddharth Dangi}, {et~al\mbox{.}}}
  \bibinfo{year}{2025}\natexlab{}.
\newblock \showarticletitle{LinkedIn Post Embeddings: Industrial Scale
  Embedding Generation and Usage across LinkedIn}. In
  \bibinfo{booktitle}{\emph{Proceedings of the 34th ACM International
  Conference on Information and Knowledge Management}}.
  \bibinfo{pages}{6038--6044}.
\newblock


\bibitem[Su et~al\mbox{.}(2023)]%
        {su2023sparcode}
\bibfield{author}{\bibinfo{person}{Liangcai Su}, \bibinfo{person}{Fan Yan},
  \bibinfo{person}{Jieming Zhu}, \bibinfo{person}{Xi Xiao},
  \bibinfo{person}{Haoyi Duan}, \bibinfo{person}{Zhou Zhao},
  \bibinfo{person}{Zhenhua Dong}, {and} \bibinfo{person}{Ruiming Tang}.}
  \bibinfo{year}{2023}\natexlab{}.
\newblock \showarticletitle{Beyond Two-Tower Matching: Learning Sparse
  Retrievable Cross-Interactions for Recommendation}. In
  \bibinfo{booktitle}{\emph{Proceedings of the 46th International ACM SIGIR
  Conference on Research and Development in Information Retrieval (SIGIR
  '23)}}. \bibinfo{publisher}{ACM}, \bibinfo{address}{New York, NY, USA},
  \bibinfo{pages}{548--557}.
\newblock
\href{https://doi.org/10.1145/3539618.3591643}{doi:\nolinkurl{10.1145/3539618.3591643}}


\bibitem[Xue et~al\mbox{.}(2025)]%
        {xue2025silvertorch}
\bibfield{author}{\bibinfo{person}{Bi Xue}, \bibinfo{person}{Hong Wu},
  \bibinfo{person}{Lei Chen}, \bibinfo{person}{Chao Yang},
  \bibinfo{person}{Yiming Ma}, \bibinfo{person}{Fei Ding},
  \bibinfo{person}{Zhen Wang}, {et~al\mbox{.}}}
  \bibinfo{year}{2025}\natexlab{}.
\newblock \bibinfo{title}{{SilverTorch}: A Unified Model-based System to
  Democratize Large-Scale Recommendation on {GPUs}}.
\newblock
\showeprint[arxiv]{2511.14881}


\bibitem[Yang et~al\mbox{.}(2025)]%
        {yang2025hit}
\bibfield{author}{\bibinfo{person}{Haoqiang Yang}, \bibinfo{person}{Congde
  Yuan}, \bibinfo{person}{Kun Bai}, \bibinfo{person}{Mengzhuo Guo},
  \bibinfo{person}{Wei Yang}, {and} \bibinfo{person}{Chao Zhou}.}
  \bibinfo{year}{2025}\natexlab{}.
\newblock \showarticletitle{HIT Model: A Hierarchical Interaction-Enhanced
  Two-Tower Model for Pre-Ranking Systems}. In
  \bibinfo{booktitle}{\emph{Proceedings of the 34th ACM International
  Conference on Information and Knowledge Management (CIKM '25)}}.
  \bibinfo{publisher}{ACM}, \bibinfo{address}{New York, NY, USA}.
\newblock
\href{https://doi.org/10.1145/3746252.3761501}{doi:\nolinkurl{10.1145/3746252.3761501}}


\bibitem[Zhai et~al\mbox{.}(2023)]%
        {zhai2023revisiting}
\bibfield{author}{\bibinfo{person}{Jiaqi Zhai}, \bibinfo{person}{Zhaojie Gong},
  \bibinfo{person}{Yueming Wang}, \bibinfo{person}{Xiao Sun},
  \bibinfo{person}{Zheng Yan}, \bibinfo{person}{Fu Li}, {and}
  \bibinfo{person}{Xing Liu}.} \bibinfo{year}{2023}\natexlab{}.
\newblock \showarticletitle{Revisiting Neural Retrieval on Accelerators}. In
  \bibinfo{booktitle}{\emph{Proceedings of the 29th ACM SIGKDD Conference on
  Knowledge Discovery and Data Mining (KDD '23)}}. \bibinfo{publisher}{ACM},
  \bibinfo{address}{New York, NY, USA}, \bibinfo{pages}{5520--5531}.
\newblock
\href{https://doi.org/10.1145/3580305.3599897}{doi:\nolinkurl{10.1145/3580305.3599897}}


\bibitem[Zhao et~al\mbox{.}(2019)]%
        {zhao2019recommending}
\bibfield{author}{\bibinfo{person}{Zhe Zhao}, \bibinfo{person}{Lichan Hong},
  \bibinfo{person}{Li Wei}, \bibinfo{person}{Jilin Chen},
  \bibinfo{person}{Aniruddh Nath}, \bibinfo{person}{Shawn Andrews},
  \bibinfo{person}{Aditee Kumthekar}, \bibinfo{person}{Maheswaran
  Sathiamoorthy}, \bibinfo{person}{Xinyang Yi}, {and} \bibinfo{person}{Ed
  Chi}.} \bibinfo{year}{2019}\natexlab{}.
\newblock \showarticletitle{Recommending what video to watch next: a multitask
  ranking system}. In \bibinfo{booktitle}{\emph{Proceedings of the 13th ACM
  conference on recommender systems}}. \bibinfo{pages}{43--51}.
\newblock


\end{thebibliography}


\end{document}